\documentclass{article}

\usepackage[final]{corl_2024} 

\usepackage{color}
\usepackage{amsfonts} 
\usepackage{textcomp}
\usepackage{pifont}
\usepackage{enumitem,amssymb}
\usepackage[pdftex]{graphicx}
\graphicspath{{figures/}{../jpeg/}}
\DeclareGraphicsExtensions{.pdf,.jpeg,.png}

\usepackage{amsmath}
\usepackage{esvect}
\usepackage{mathtools}
\usepackage[]{hyperref}
\usepackage{multirow}
\usepackage{rotating}

\title{DOFS: A Real-world 3D \underline{D}eformable \underline{O}bject Dataset with \underline{F}ull \underline{S}patial Information for Dynamics Model Learning}
\author{
  Zhen Zhang, Xiangyu Chu, Yunxi Tang, K. W. Samuel Au\\
  Department of Mechanical and Automation Engineering\\
  The Chinese University of Hong Kong, HKSAR\\
  Multi-scale Medical Robotics Center \\
  \texttt{zhzhen@link.cuhk.edu.hk} 
}

\begin{document}
\maketitle


\begin{abstract}
This work proposes DOFS, a pilot dataset of 3D deformable objects (DOs) (e.g., elasto-plastic objects) with full spatial information (i.e., top, side, and bottom information) using a novel and low-cost data collection platform with a transparent operating plane. The dataset consists of active manipulation action, multi-view RGB-D images, well-registered point clouds, 3D deformed mesh, and 3D occupancy with semantics, using a pinching strategy with a two-parallel-finger gripper. In addition, we trained a neural network with the down-sampled 3D occupancy and action as input to model the dynamics of an elasto-plastic object. Our dataset and all CADs of the data collection system will be released soon on our  \href{https://tmmdhz.github.io/DOFS.github.io/}{website}.

\end{abstract}



\section{Introduction}
\begin{figure}[htbp]
    \centering
    \includegraphics[width=1.0\linewidth]{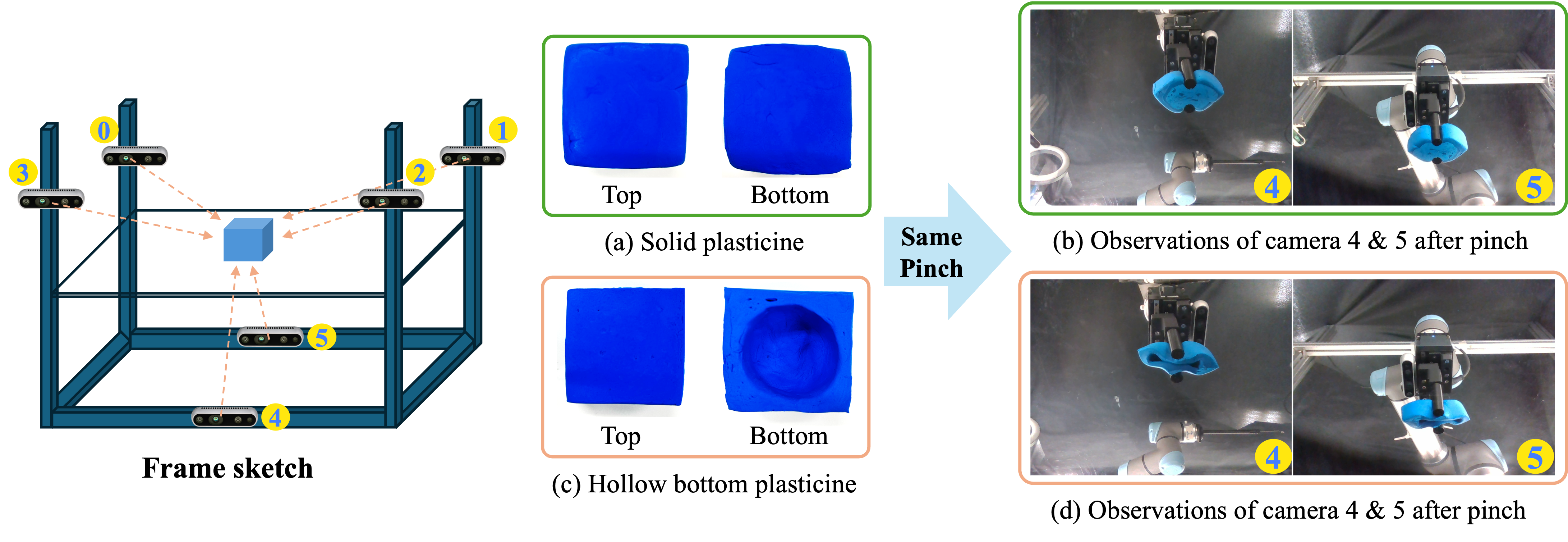}
    \caption{\textbf{Dynamics difference caused by the hollow bottom}. For two pieces of plasticine of the same appearance and size, one is solid and another is hollow, the deformation results are completely different with the same action.}
    \label{bottom}
\end{figure}

Robot manipulation of 3D Deformable Objects is essential for many activities and applications in the real world, such as household~\cite{foldreal,speedfolding} and healthcare~\cite{Scheikl_2023}, and is still an open challenge despite extensive studies. Recently, data-driven solutions have shown impressive and promising results in 3D deformable object manipulation by learning-based approaches~\cite{learningpred,robocook}, where sufficient data is essential to improve model training or policy learning. To obtain training data, some previous works collected synthetic data from simulators~\cite{DefGraspSim}. Still, there is an unavoidable gap between the real world and the simulator since the existing simulators cannot accurately simulate all real-world physical characteristics (e.g., friction, impact, and stiffness)~\cite{Sim2Real}. To mitigate the gap, some researchers ~\cite{sculptbot,deformnet,robocraft,pokeflex} collect Real-World Data (RWD); for example,~\cite{sculptbot,deformnet} collects RGB-D images and point clouds,~\cite{robocraft} collects 3D mesh models,~\cite{pokeflex} uses a professional system with 106 cameras to obtain the 3D reconstructions of deformed mesh. However, there are still some limitations. The data of~\cite{sculptbot, deformnet, robocraft, pokeflex} are not full-spatial without considering the bottom side of 3D DOs as they are placed on the operating plane, which may lead to the dramatic dynamics difference caused by the hollow bottom as shown in Fig.~\ref{bottom}. Besides,~\cite{pokeflex} is highly costly and complicated to replicate.

We propose a low-cost and novel data collection platform to collect the dataset of 3D elasto-plastic object manipulation with full spatial information. Our DOFS dataset includes  3D reconstructions of deformed meshes,  corresponding actions,  well-registered 3D point clouds, and multi-view RGB-D images. In addition, we also provide the 3D occupancy of the manipulation scenario, which is a very dense, high-resolution, and semantic-rich representation. 


\section{Methods}
\label{methods}
To build our DOFS dataset, we first built a novel data collection platform including an aluminum frame with a transparent operating plane, a robotic arm, and 6 RGB-D cameras. We proposed a generalizable pipeline to process collected raw data and generate the structured dataset. Finally, we trained a dynamics model of the plasticine using our DOFS dataset to validate the quality of the dataset.

\subsection{Hardware Setup}
\begin{figure}[htbp]
    \centering
    \includegraphics[width=0.8\textwidth]{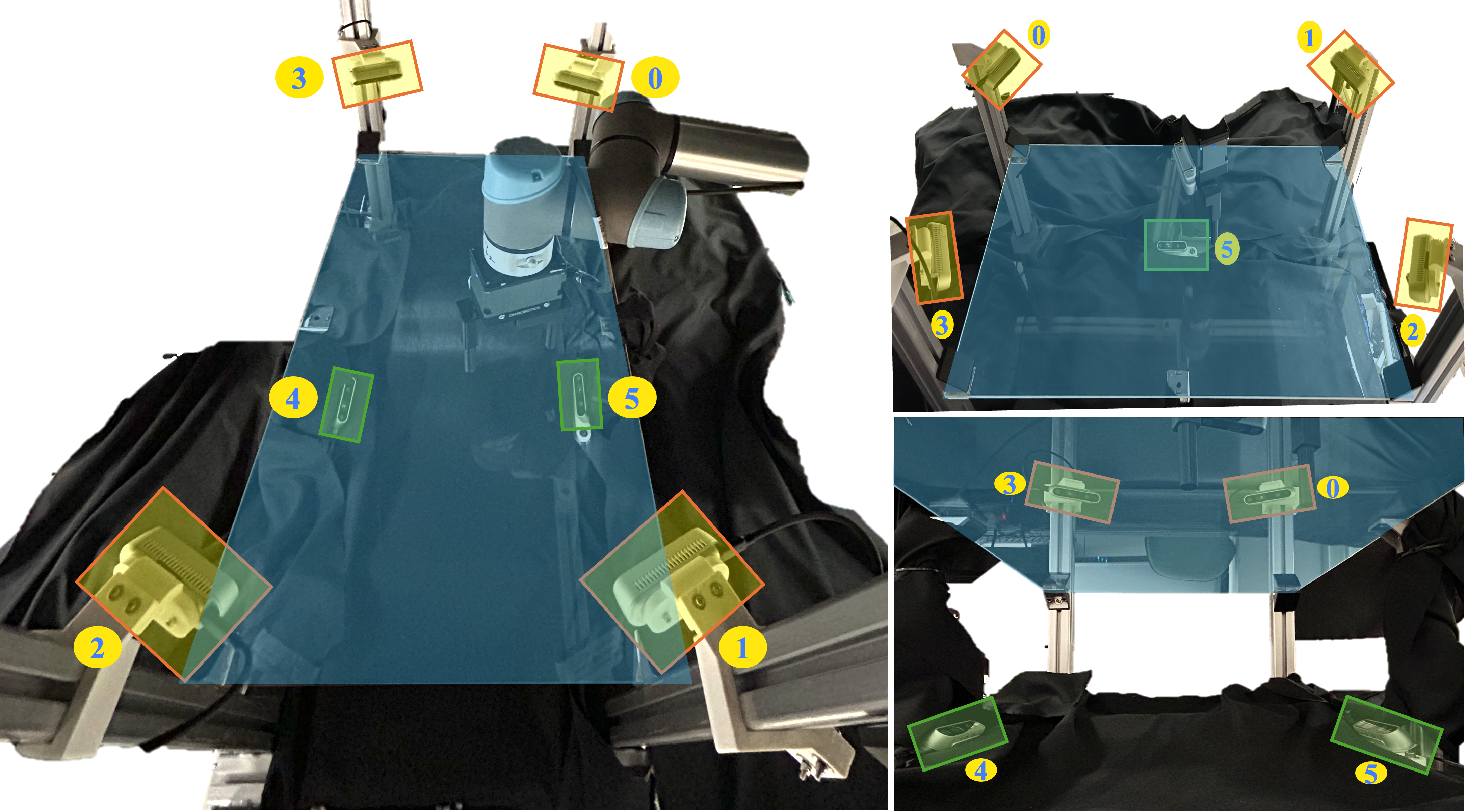}
    \caption{\textbf{Hardware setup of our data collection platform}. The blue part is a transparent acrylic board that serves as an operating plane. The orange parts are 4 cameras above the plane to collect top and side information. The green parts are 2 cameras below the plane to collect bottom-side information.}
    \label{platform}
\end{figure}

In our data collection system, we propose a novel hardware setup to capture the multi-view RGB-D images from the top, sides, and bottom of objects for full spatial information collection. To be specific,  we used aluminum profiles to build a frame and installed a transparent acrylic board in the middle of the frame as the operating plane, where the objects are placed. We use 6 Intel RealSense D435i cameras and install 4 cameras on top of it and 2 cameras on the bottom from different positions and angles respectively to collect multi-view RGB-D images of the manipulation scenario to obtain dense point clouds, as shown in Fig.~\ref{platform}.

\begin{figure}[htbp]
    \centering
    \includegraphics[width=0.9\linewidth]{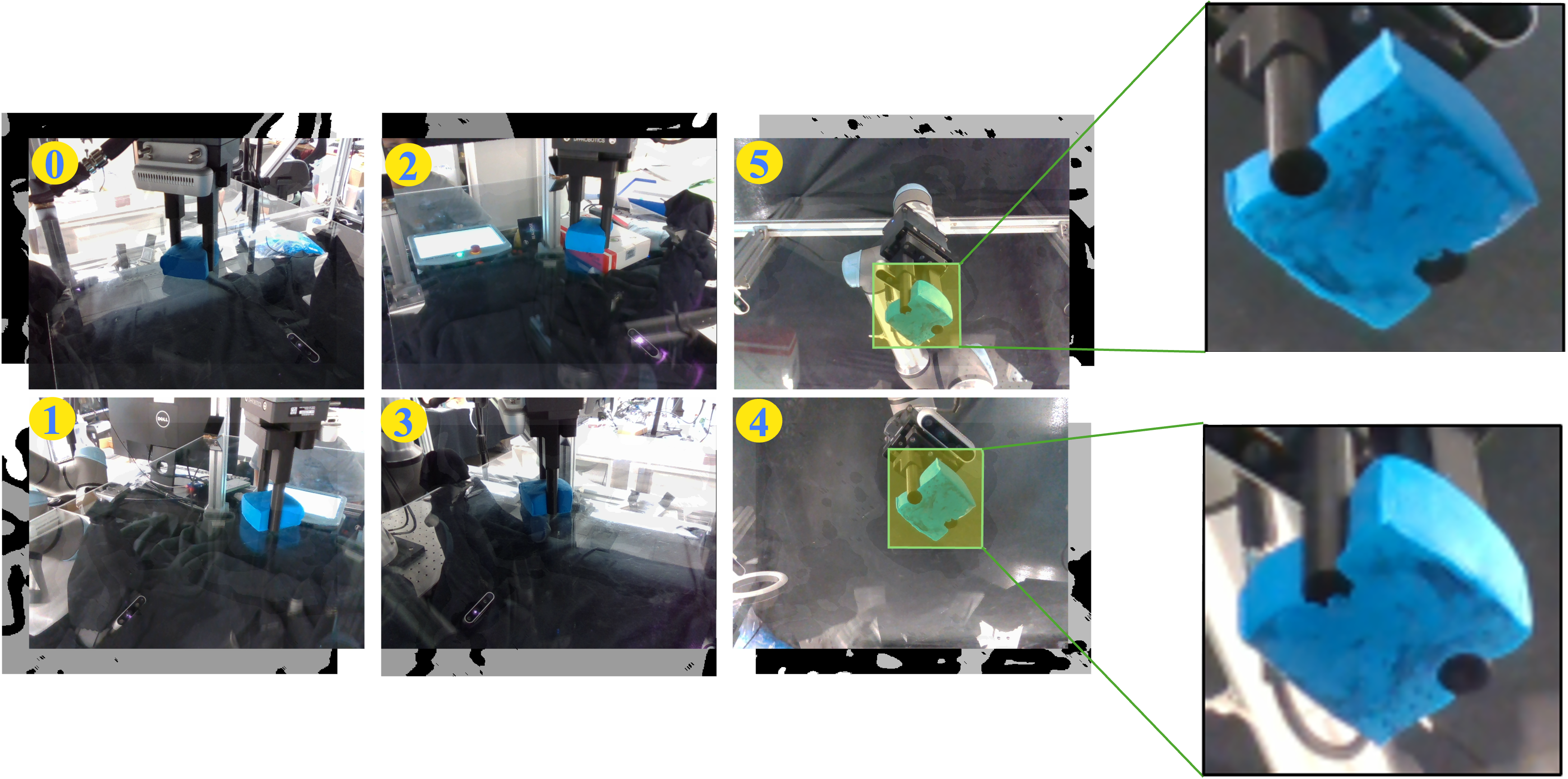}
    \caption{\textbf{Six RGB-D images captured from RealSense D435i}. Four images from the cameras installed above the operating plane show the top and side information of the plasticine. Two images from the cameras installed below the operating plane show the bottom side information of the plasticine.}
    \label{datasets}
\end{figure}

\subsection{Data Collection and Processing}	
We use ROS~\cite{ROS} to synchronize and record the actions of the robotic arm and 6 RGB-D cameras during manipulation. We use Iterative Closest Point (ICP)~\cite{ICP} to register the point cloud generated from 6 RGB-D cameras to obtain the full spatial information. After that, we used geometry information and registered point clouds to reconstruct the deformed meshes. Moreover, we selected the voxel size $[0.002,0.002, 0.002]m$ to generate the 3D occupancy data and assign the semantic label for the operating plane, two fingers, and deformable objects.


\section{Results}
\label{sec:result}

\begin{figure}[htbp]
    \centering
    \includegraphics[width=0.8\linewidth]{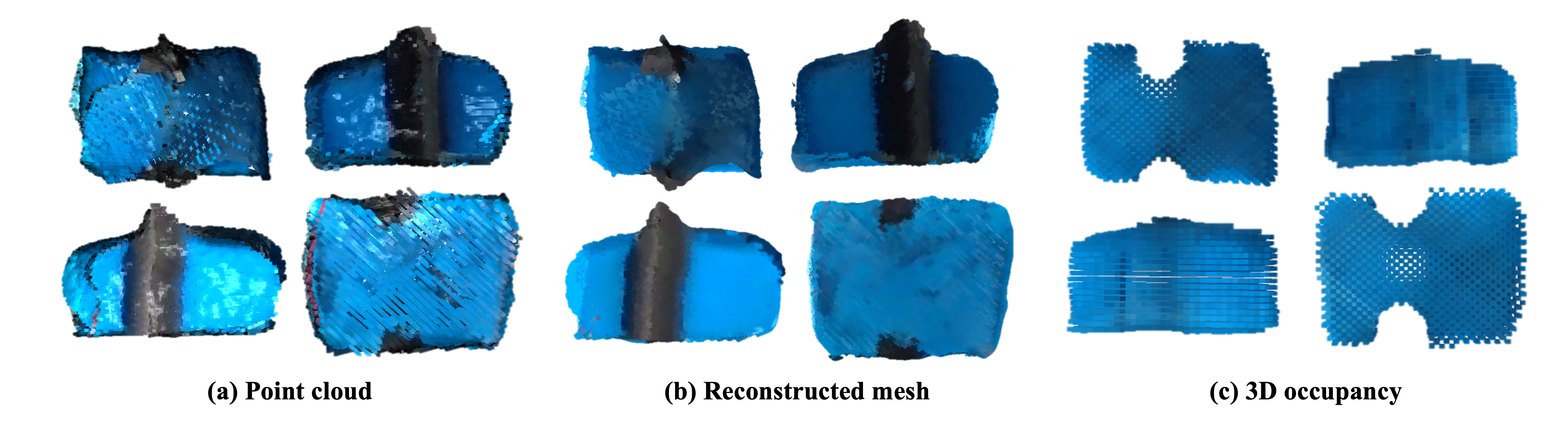}
    \caption{\textbf{Data visualization of one frame during manipulation.} (a): Well-registered point cloud of pinched plasticine without background. (b): Reconstructed deformed mesh. (c): 3D occupancy of plasticine.}
    \label{deformation exp}
\end{figure}

We use 6 well-calibrated cameras to collect 6 RGB-D images shown in Fig.~\ref{datasets} and record the action (i.e., the 3D positions of two fingers) simultaneously during manipulation. We use 6 RGB-D images and their intrinsic and extrinsic matrix to generate dense point clouds shown in Fig.~\ref{deformation exp} (a) for deformed mesh reconstruction shown in Fig.~\ref{deformation exp} (b). Specifically, We use UR5e with a two-finger paralleled gripper of DH Robotics PGI-140-80 to manipulate the plasticine. We discrete one pinch action into 30 steps to pinch deformable objects step by step. Our current DOFS dataset contains 150 pinches(i.e., 4500 frames), and one frame captures the object deformation shown in Figure \ref{deformation exp} (c). In addition, each frame contains the following information:

\begin{itemize}
  \item 3D deformed mesh of the plasticine
  \item 3D positions of two fingers
  \item Multi-view RGB-D images from 6 cameras
  \item Registered 3D point clouds
  \item 3D occupancy with semantics
\end{itemize}

\begin{figure}[htbp]
    \centering
    \includegraphics[width=1.0\linewidth]{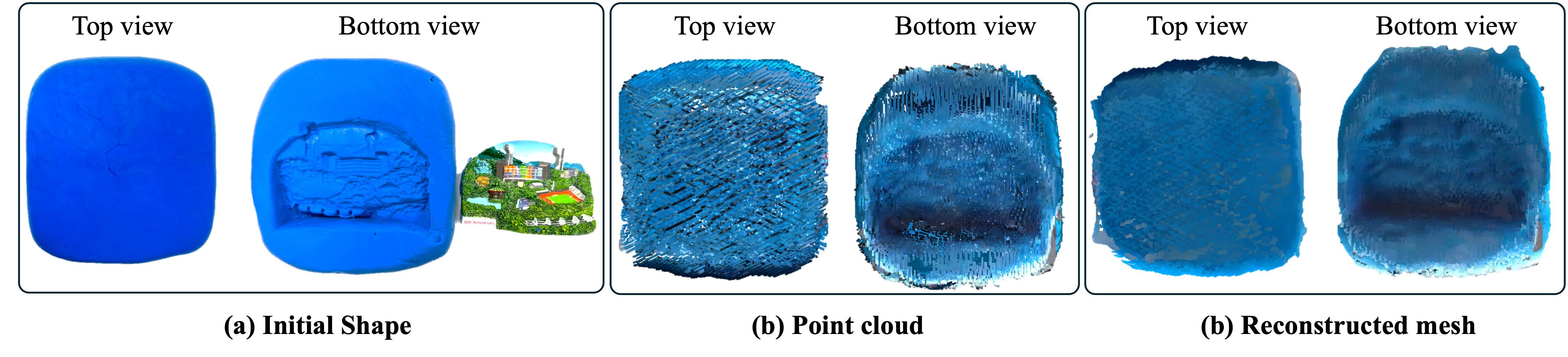}
    \caption{\textbf{Visualization of Full spatial information collection}. (a): We embed the cartoon model into the bottom of the plasticine and take it out. (b): The visualization of the point cloud of plasticine. (c): The deformed mesh of plasticine.}
    \label{fullspatial}
\end{figure}

We embedded a cartoon model into the bottom of the plasticine and took it out, shown in Fig.~\ref{fullspatial} (a). We capture the data of the plasticine placed on the operating plane and visualize the point cloud and deformed mesh with geometry details shown in Fig.~\ref{fullspatial} (b) and Fig.~\ref{fullspatial} (c), respectively, for reporting full-spatial results.

\begin{figure}[htbp]
    \centering
    \includegraphics[width=0.7\linewidth]{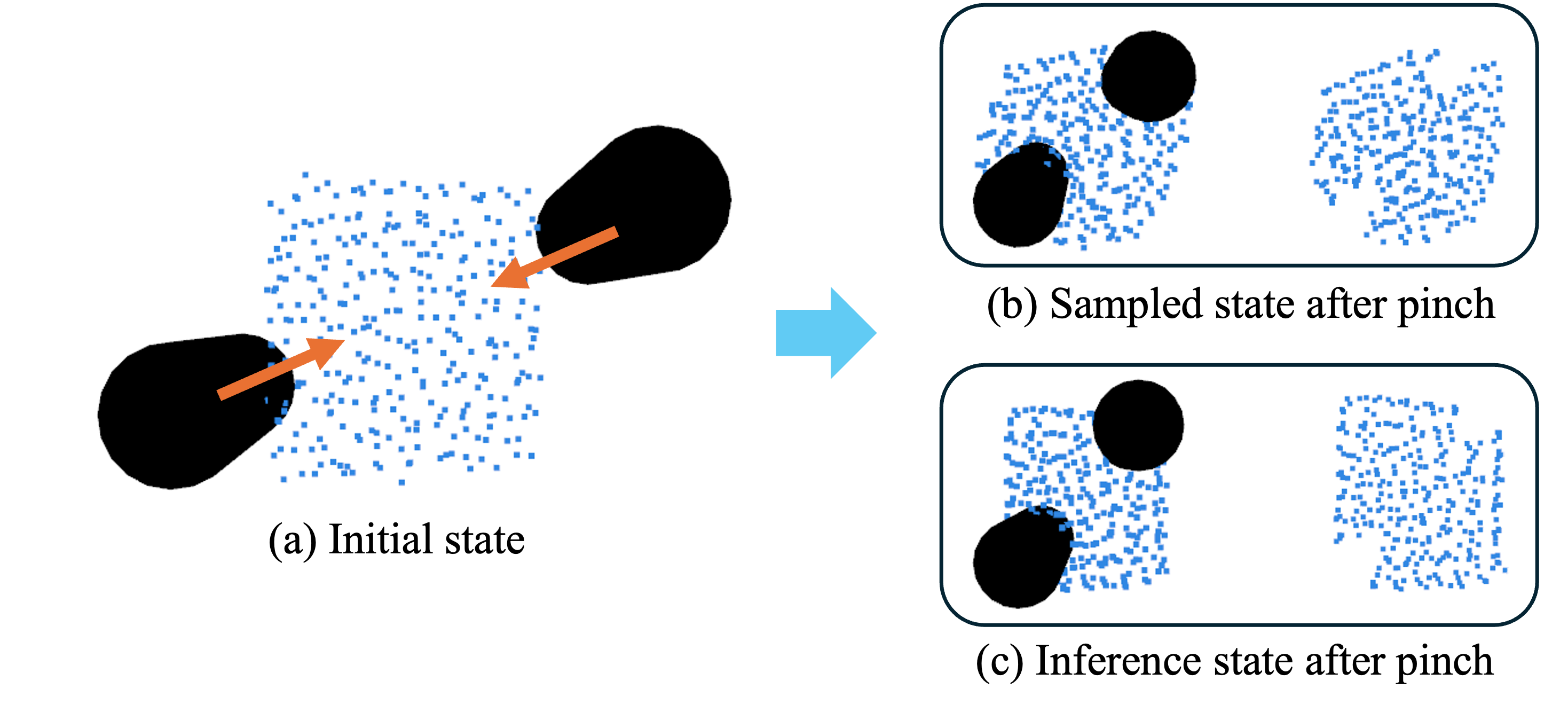}
    \caption{\textbf{State transformation after pinching the plasticine}. (a): Visualization of the initial state of the plasticine and two fingers, the yellow arrow indicates the pinch direction. (b): The ground-truth state after the pinch. (c): The inference state of the learned dynamics model.}
    \label{results}
\end{figure}

To evaluate the quality of our DOFS dataset, we downsample the generated 3D occupancy and use it to train a dynamics model of the plasticine. We trained it for 8 epochs using a machine with Intel 12700K x 12 Core Processor CPU and NVIDIA GeForce RTX 4080 GPU (16GB memory). The preliminary experiment shown in Fig.~\ref{results} shows that our model can predict the state of plasticine after an action is applied.

\section{Conclusion}
\label{sec:conclusion}
Our low-cost and novel data collection platform can effectively generate detailed 3D deformed meshes and dense, high-resolution 3D occupancy data. The learned dynamics model shows great potential for deformable object manipulation. To promote our DOFS dataset and system, we will choose 3D DOs with different physical properties (e.g., stiffness) and geometry shapes to diversify our dataset. In addition, we will change the gripper shape to apply other manipulation strategies (e.g., cut, roll). We consider that our DOFS dataset and data collection system have the potential to advance the research on robotic fine and dexterous manipulation, and we look forward to sharing this DOFS dataset and platform available with the community.


\clearpage
\acknowledgments{
This work was supported by the Research Grants Council (RGC) of Hong Kong under Grant 14209719. We thank my colleagues, Lam Him Kwok and Yongxuan Feng, for their assistance with the hardware setup.
}


\bibliography{bibliography.bib}

\end{document}